\UseRawInputEncoding
\pdfoutput=1
\documentclass{article}

     \usepackage[nonatbib, preprint]{neurips_2023}

\usepackage[utf8]{inputenc} 
\usepackage[T1]{fontenc}    
\usepackage{hyperref}       
\usepackage{url}            
\usepackage{booktabs}       
\usepackage{mathtools,amssymb}
\usepackage{amsfonts}       
\usepackage{nicefrac}       
\usepackage{microtype}      
\usepackage{pgfplots,pgfplotstable}
\pgfplotsset{compat=1.14}
\usepackage{array,colortbl}
\usepackage{xcolor}
\usepackage{algorithm,algorithmicx,algpseudocode}
\usepackage[capitalise]{cleveref}
\usepackage{caption}
\usepackage{graphbox}
\usepackage{placeins}
\usepackage{wrapfig}
\usepackage{graphicx}
\usepackage{subcaption}
\usepackage{etoolbox}
\usepackage[numbers,square,sort&compress]{natbib}
\usepackage{overpic}
\usepackage{multirow}
\usepackage{dashrule}

\newtoggle{hqfigures}
\toggletrue{hqfigures}

\newcommand{\bzero}{\mathbf{0}}

\newcommand{\tabincell}[2]{\begin{tabular}{@{}#1@{}}#2\end{tabular}}
\newcommand{\ourmodel}{CamoDiffusion}

\title{\ourmodel: Camouflaged Object Detection via Conditional Diffusion Models}

\author{%
  Zhongxi Chen$^{1,2}$, Ke Sun$^{2,3}$, Xianming Lin$^2$\thanks{Corresponding Author.}, Rongrong Ji$^{1,2}$ \\
  $^1$Institute of Artificial Intelligence, Xiamen University \\
  $^2$Media Analytics and Computing Lab, Department of Artificial Intelligence\\
  $^3$School of Informatics, Xiamen University, 361005, China \\
  {\tt\small https://github.com/Rapisurazurite/CamoDiffusion} \\
}

\begin{document}

\maketitle

\begin{abstract}
  Camouflaged Object Detection (COD) is a challenging task in computer vision due to the high similarity between camouflaged objects and their surroundings. 
  Existing COD methods primarily employ semantic segmentation, which suffers from overconfident incorrect predictions. 
  In this paper, we propose a new paradigm that treats COD as a conditional mask-generation task leveraging diffusion models. Our method, dubbed \ourmodel, employs the denoising process of diffusion models to iteratively reduce the noise of the mask.
  Due to the stochastic sampling process of diffusion, our model is capable of sampling multiple possible predictions from the mask distribution, avoiding the problem of overconfident point estimation. 
  Moreover, we develop specialized learning strategies that include an innovative ensemble approach for generating robust predictions and tailored forward diffusion methods for efficient training, specifically for the COD task.
  Extensive experiments on three COD datasets attest the superior performance of our model compared to existing state-of-the-art methods, particularly on the most challenging COD10K dataset, where our approach achieves \textbf{0.019} in terms of MAE.
\end{abstract}

\section{Introduction}
Camouflage, a pervasive defense strategy in nature, endows organisms with the capacity to meld seamlessly into their surroundings, thereby allowing them to elude predators or approach prey surreptitiously~\cite{fangtpami2021}. As a result, Camouflaged Object Detection (COD) has materialized as a rapidly expanding research field, focusing on the identification of hidden objects or organisms in their natural habitats. This area boasts applications across a multitude of sectors, including species conservation~\cite{nafus2015hiding, perez2012early}, medical image segmentation~\cite{dong2021PolypPVT, fang20PraNet, ji2022vps, fan2020infnet}, and industrial flaw detection~\cite{bhajantri2006camouflage}. Notwithstanding considerable research attention, the discernment of camouflaged objects continues to pose significant challenges due to the striking resemblance they share with their environments. Consequently, there is an exigent demand for innovative and refined COD techniques.

To address the camouﬂaged properties of the foreground object, numerous strategies have been proposed for this task, predominantly encompassing three perspectives~\cite{fan2023advances}: 1) Multi-stream frameworks~\cite{yan2021mirrornet, wu2023source, kajiura2021improving, zhong2022detecting, pang2022zoom, zheng2023mffn}, which exploit multiple input streams to explicitly learn multi-source representations. 2) Bottom-up and top-down frameworks~\cite{fan2020camouflaged, fangtpami2021, mei2021camouflaged, ren2023tanet, hu2023high, huang2023feature, yin2023camoformer} that harness deeper features to progressively enhance shallower ones in a single feed-forward pass. 3) Branched frameworks~\cite{lv2021simultaneously, ji2023gradient, lv2023towards, li2021uncertainty, zhai2023mglv2, yang2021uncertainty, zhu2022can, zhou2022feature, sun2022boundary, zhai2021mutual}, constituting a single-input multiple-output architecture, comprising both segmentation and auxiliary task branches.
Such techniques mainly build upon the foundational semantic segmentation paradigm, which capitalizes on a learning-based backbone for feature extraction, subsequently employing a decoder head to yield a segmentation mask. However, such a paradigm is prone to confuse the subtle deviation between boundaries and surroundings of camouﬂaged objects due to the body outline disguising~\cite{sun2022boundary}, leading to unsatisfactory predictions. Furthermore, the segmentation paradigm depends on pixel-wise probabilities leading to overconfident incorrect predictions.

Driven by the above challenges, we introduce a new paradigm that treats COD as a conditional mask-generation task based on diffusion models~\cite{ho2020denoising, song2020score, song2019generative}. 
As the diffusion models employ a step-by-step denoising method for generating results, they possess impressive generative capabilities with a good awareness of conditions. 
Following this paradigm, we propose a new framework named \ourmodel, which utilizes the denoising process of the diffusion models to gradually remove the bias between the initial noise and ground truth in an iterative manner and incorporate the image as the auxiliary condition. 
Specifically, we propose an Adaptive Transformer Conditional Network and Denoising Network to encode the condition image and denoise the mask, respectively. Furthermore, we design specialized learning strategies during the training and sampling process to ensure efficient training and achieve more robust results, including an SNR-based variance schedule, Structure Corruption, and Consensus Time Ensemble.
Compared to previous COD methods, our method has the following advantages: 1) \ourmodel\ can significantly enhance the capacity to handle subtle details of the foreground. 2) \ourmodel\ allows for a more comprehensive understanding of the intrinsic differences between camouflaged objects and their surroundings, leading to better generalization and less mis-segmentation.
3) \ourmodel\ can sample multiple possible predictions, avoiding the problem of overconfident point estimation.

Our main contributions can be summarized as follows:
\begin{enumerate}
\item We are the first to treat the COD task as a mask generation paradigm and use a conditional diffusion framework to form predictions.
\item We propose a novel and effective framework called \ourmodel, which uses a specially designed network structure and learning strategies to generate more accurate and generalized results for the camouflaged object detection task.
\item Our \ourmodel\ achieves state-of-the-art (SOTA) performance on three COD datasets, demonstrating its superior effectiveness in completing the COD task.
\end{enumerate}


\section{Related Work}
\subsection{Camouflaged Object Detection}

In recent years, CNN-based methods have achieved remarkable advancements in the COD task by employing intricate strategies. For instance, SINet-V2~\cite{fan2022concealed} and BASNet~\cite{qin2019basnet} utilize coarse-to-fine strategies that initially predict a rough segmentation of the entire region and subsequently refine it using various techniques. Multi-task learning-based approaches, such as LSR~\cite{lv2021simultaneously}, BGNet~\cite{sun2022boundary}, and DGNet~\cite{ji2023gradient}, incorporate common detection tasks like ranking, localization, edge detection, and gradient generation to assist the binary segmentation task and enhance COD performance. ZoomNet~\cite{pang2022zoom} and MFFN~\cite{zheng2023mffn} employ multiple augmented views as inputs to extract features at different scales or angles and fuse them to obtain an improved feature map. On the other hand, Vision Transformer has recently demonstrated exceptional performance in COD by using self-attention mechanisms to model long-range dependencies. Notable works in this area include DTINet~\cite{liu2022boosting}, FSPNet~\cite{huang2023feature}, and HitNet~\cite{hu2023high}, which adopt a dual-task interactive Transformer, non-local mechanisms, and iterative refinement of low-resolution representations through feedback, respectively.
While recent years have seen substantial progress in COD models, most still suffer from the limitation of providing only point estimation. To address this, we generate a distribution of segmentation masks by leveraging the inherent stochastic sampling process of diffusion.



\subsection{Diffusion Models for Image Segmentation}
Diffusion models~\cite{ho2020denoising, song2020score, song2019generative} are parameterized Markov chains that gradually denoise data samples starting from random noise. 
Although originally applied in fields without definitive ground truth~\cite{zhang2023adding, dhariwal2021diffusion}, recent research has demonstrated their effectiveness for problems with unique ground truths, such as super-resolution~\cite{li2022srdiff, wang2021s3rp}, deblurring~\cite{whang2022deblurring, lee2022progressive}, and image segmentation~\cite{baranchuklabel, amit2021segdiff, chen2022generalist}. 
This work highlights the potential of diffusion models in various segmentation tasks, such as remote sensing change detection~\cite{bandara2022ddpm} and medical image segmentation~\cite{wolleb2022diffusion, wu2022medsegdiff, rahman2023ambiguous}.
For instance, \citet{wolleb2022diffusion} used a diffusion model to perform lesion segmentation with an image as a prior, while \citet{rahman2023ambiguous} investigated the ability of diffusion models to perform ambiguous image segmentation. 
However, the current diffusion models rely on a U-Net~\cite{ronneberger2015u} based on CNN with limited input resolution, which can not provide the necessary global information for complex tasks such as COD. 

\section{Method}
\begin{figure}
  \centering
  \includegraphics[width=1.0 \textwidth, trim=50 200 40 240,clip]{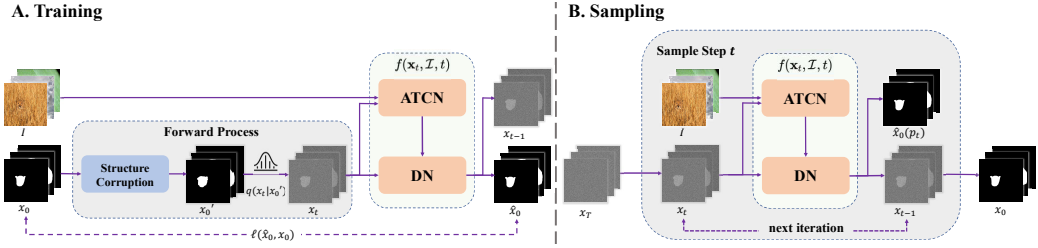}  
  \small
  \vspace{-5mm}
  \caption{The overall framework of our \ourmodel. Our model $f(\mathbf x_t, \mathcal I, t)$ includes an Adaptive Transformer Conditional Network (ATCN), and a Denoising Network (DN). In the Training stage, the mask $\mathbf x_0$ is subject to Structure Corruption and Gaussian noise to generate the noised mask $\mathbf x_t$ at time $t$. Subsequently, our model predicts the denoised mask $\hat{\mathbf x}_0$ and is trained by minimizing the loss between the prediction and the ground truth. During the Sampling procedure, our model denoises the random noise $\mathbf x_T$ for $T$ steps to produce the final prediction $\mathbf x_0$.}
  \label{fig:OverallFramework}
  \vspace{-3mm}
\end{figure}
In this section, we introduce our \ourmodel\ framework, which progressively generates predictions by conditioning each subsequent step with the image prior. As depicted in Fig.~\ref{fig:OverallFramework}, 
the CamoDiffusion framework comprises an Adaptive Transformer Conditional Network (ATCN) and a fundamental Denoising Network (DN). The ATCN is employed to extract image features and encode underlying conditions, whereas the DN serves to eliminate noise and formulate predictions. 
In the Training stage (Fig.~\ref{fig:OverallFramework} A), our model is optimized to denoise noisy masks that are corrupted using a specially designed forward process by learning the reverse distribution. During the Sampling stage (Fig.~\ref{fig:OverallFramework} B), the sampled noise undergoes progressive denoising, culminating in the final prediction. We first introduce the basic background and notation in Section~\ref{sec:BN}. Then we discuss the architectural details of the ATCN and DN in Section~\ref{sec:AD}. The details of the specially designed learning strategies are mentioned in Sec.~\ref{sec:TS} and Sec.~\ref{sec:SS}.

\subsection{Background and Notation}
\label{sec:BN}
Our \ourmodel\ is based on diffusion models, which include a forward process, where a mask is gradually noised, and a reverse process, where noise is transformed back to the target distribution. 
Given a training sample $\mathbf x_0 \sim q(\mathbf x_0)$, the noised version $\{\mathbf x_t\}_{t=1}^T$ are obtained according to the following Markov process:
\begin{equation}
    q ( \mathbf x _ { t } | \mathbf  x _ { t-1} ) = \mathcal N ( \mathbf  x _ { t } ; \sqrt { 1 - \beta _ { t } } \mathbf  x _ { t-1 } , \beta _ { t }  \textbf I ) ,
\end{equation}
where $t$ runs from $1$ to $T$, and variance is controlled by noise schedule $\beta_t \in (0,1)$. The marginal distribution of $\mathbf x_t$ can be described as:
\begin{equation}
    q(\mathbf x_t|\mathbf x_0) = \mathcal N(\mathbf x_t;\sqrt{\bar\alpha_t} \mathbf x_0,(1-\bar \alpha_t) \textbf I),
\end{equation}
where $\bar \alpha_t=\prod_{i=1}^T\alpha_t, \alpha_t = 1-\beta_t$. 
Starting from $p(\mathbf x_T)=\mathcal N(\mathbf x_T;\bzero,\textbf I)$, the reverse process uses a neural network $f_\theta$ to create a sequence of incremental denoising operations to obtain back the clean mask. The network learns the reverse distribution:
\begin{equation}
    p(\mathbf x_{t-1}|\mathbf x_t):= \mathcal N(\mathbf x_{t-1} ; \mu_\theta(\mathbf x_t,t),\Sigma_\theta(\mathbf x_t,t)).
\end{equation}
While it is feasible to model $\mathbf x_{t-1}$ directly, \citet{ho2020denoising} suggested that a consistent output space for the network leads to enhanced performance. In our proposed \ourmodel, we choose to train a network $f_\theta(\mathbf x_t, \mathcal I, t)$ to predict the denoised mask $\hat {\mathbf x}_0$ conditional on image $\mathcal I$. In practice, $\Sigma_\theta(\mathbf x_t, t)$ is set to $\sigma_t^2=\frac{1-\bar \alpha_{t-1}}{1-\bar \alpha_t}\beta_t$, and $\mu_\theta(\mathbf x_t,t)$ can be expressed as: 
\begin{equation}
    \mu_\theta(\mathbf x_t,t) = \frac{\sqrt{\alpha_t}(1 - \bar{\alpha}_{t-1})}{1 - \bar{\alpha}_t} \mathbf{x}_t + \frac{\sqrt{\bar{\alpha}_{t-1}}\beta_t}{1 - \bar{\alpha}_t} \hat{\mathbf{x}}_0,
\label{con:mu_theta}
\end{equation}
where $\hat{\mathbf{x}}_0$ is predicted by our model $f_\theta(\mathbf x_t, I, t)$, which consists of ATCN and DN.
Then we optimize $\ell(\hat{\mathbf x}_0,\mathbf x_0)$ to train our model, Formally,
\begin{equation}
\ell(\hat{\mathbf x}_0, \mathbf x_0) = \ell^w_{IoU}(\hat{\mathbf x}_0, \mathbf x_0) + \ell^w_{BCE}(\hat{\mathbf x}_0, \mathbf x_0),
\end{equation}
where $\ell^w_{IoU}$ and $\ell^w_{BCE}$ represent weighted intersection-over-union (IoU) loss and weighted binary cross entropy (BCE) loss.

\subsection{Architecture Design}
\label{sec:AD}

\begin{figure}
  \centering
  \includegraphics[width=1.0 \textwidth, trim=0 125 0 15,clip]{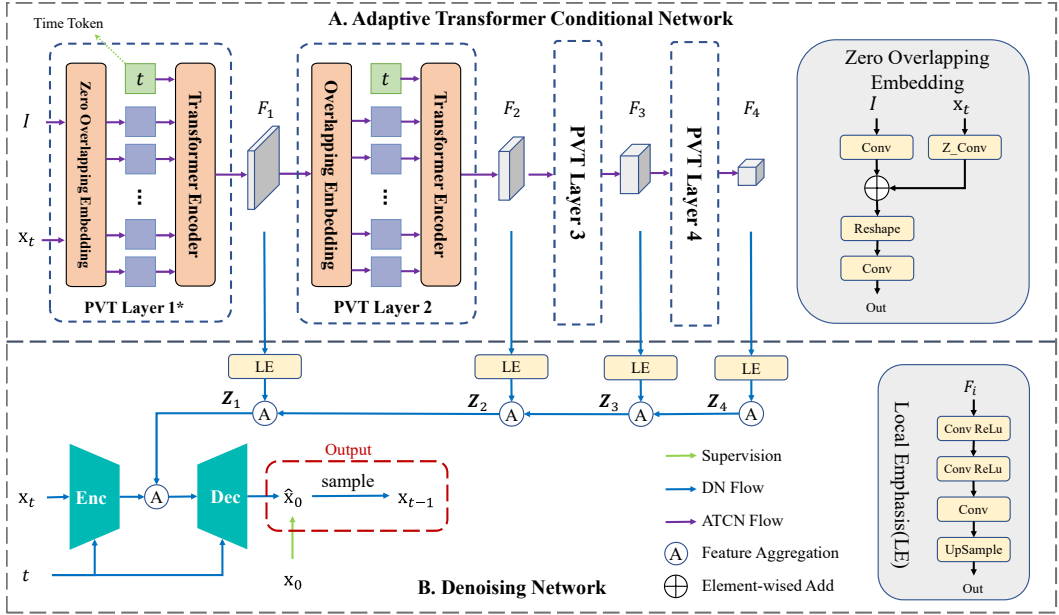}
  \small
  \caption{The architecture design of our \ourmodel. It consists of an Adaptive Transformer Conditional Network for extracting multi-scale features as conditions, and a Denoising Network for recovering clear mask predictions from the noised mask. "*" represents that the PVT layer is modified by our proposed ZOE.}
  \label{fig:ModelArchitecture}
  \vspace{-3mm}
\end{figure}

In contrast to the conventional segmentation paradigm, our proposed model employs conditional diffusion models to generate predictions, thus deviating significantly from the architectural design of previous models. Specifically, as illustrated in Fig.~\ref{fig:ModelArchitecture}, we utilize a transformer-based network to extract hierarchical image features as conditions, which are then integrated with the downstream denoising network. 
We discuss the design details of these networks in the following sections.

\paragraph{Adaptive Transformer Conditional Network.}
Within our framework, the role of the conditional network is to enable the downstream denoising network at each step to adequately discern the camouflaged image, thereby distinguishing the camouflaged target. We identify two primary challenges in the design of a conditional network: 1) Extracting more discriminative image features; 2) Adaptively supplying conditional features in accordance with the denoising step. 

To address the aforementioned challenges, we propose a specially designed Adaptive Transformer Conditional Network (ATCN) to dynamically and comprehensively extract image features. Specifically, as depicted in Fig.~\ref{fig:ModelArchitecture} A, the ATCN comprises Pyramid Vision Transformer (PVT)~\cite{wang2022pvt} layers that extract multi-scale features ($\left\{F_i\right\}_{i=1}^4$) from image $\mathcal I$ using $\mathbf x_t$ and $t$. The PVT layer which consists of a Patch Embedding module and a Transformer Encoder, recognized for its potent feature extraction capabilities, in conjunction with a multi-scale mechanism, ensures a comprehensive fusion of the camouflaged image from coarse-grained to fine-grained features at each step. Moreover, while prior research~\cite{amit2021segdiff, chen2022generalist} utilized a conditional network that solely employed images as input and used the same image features in the reverse process, we maintain that the requisite features for denoising at varying stages are not uniform. The conditional network should concentrate on the overall image when the Signal-to-Noise Ratio (SNR) of the noised mask is low, and shift focus to detailed regions once the contour of the mask has been restored. Therefore, we design \textbf{Zero Overlapping Embedding} and \textbf{Time Token Concatenation} to inject the noised mask $\mathbf x_t$ and time $t$ adaptively.



\normalfont{\normalsize\bfseries \textbullet\ \ Zero Overlapping Embedding:}
To incorporate the noise mask $\mathbf x_t$ into the PVT without destroying the original transformer structure and pre-training parameters, we propose the Zero Overlapping Embedding (ZOE) in the first layer of ATCN instead of the original position encoding module Overlapping Embedding (OE). Specifically, ZOE uses an extra convolution layer initialized with zeros, which gradually introduces $\mathbf x_t$ in a controlled manner without affecting the position encoding during initialization.
The mathematical representation of the embedding of the $i$-th layer is given by:
\begin{equation}
    \mathbf{tokens} = 
    \begin{cases}
            \mathrm{LN}(\mathrm{Reshape}(\mathrm{Conv}(\mathcal I) + \mathrm{Z\_Conv}(\mathbf x_t))), & i=1(\mathrm{ZOE}), \\ 
            \mathrm{LN}(\mathrm{Reshape}(\mathrm{Conv}(F_{i-1}))), & i \neq 1(\mathrm{OE}).
    \end{cases}
\end{equation}
Here, $\mathrm{Conv}(\cdot)$ refers to a convolution layer, $\mathrm{Z\_Conv}(\cdot)$ denotes a convolution layer with weight and bias initialized as zeros, $\mathrm{Reshape}(\cdot)$ converts the feature map obtained from the convolution to token form, and $\mathrm{LN}(\cdot)$ denotes layer normalization.

This module effectively addresses the challenges associated with the conventional approach of modifying the input channel of OE, which leads to the re-initialization of layer parameters and longer convergence times. 
\normalfont{\normalsize\bfseries \textbullet\ \ Time Token Concatenation:}
Beyond the noisy mask, we anticipate our ATCN to autonomously adjust the characteristics of the condition according to the temporal stage. To incorporate such temporal information into the Transformer, we propose Time Token Concatenation. Specifically, the time token is concatenated with the patch sequence procured via OE (or ZOE) and fed into the Transformer encoder. 

\paragraph{Denoising Network}
To produce the denoised mask prediction $\hat{\mathbf x}_0$ and $\mathbf x_{t-1}$, we proposed DN (Fig. \ref{fig:ModelArchitecture} B), which utilizes a simple U-shaped structure, with the input consisting of time $t$, multi-scale feature maps $\{ F_i\}_{i=1}^4$ extracted by ATCN, and the noised mask $\mathbf x_t$. Instead of designing complex hierarchically refined decoders, we achieved satisfactory results by utilizing the simple structure due to the iterative denoising process of diffusion. Specifically, we adopt the Local Emphasis ($\mathrm{LE}$) module~\cite{wang2022stepwise} to upsample the multi-scale features $\{F_i\}_{i=1}^4$ to the same size $H/4 \times W/4$, denoted as: $F_i^{up} = \mathrm{LE}(F_i)$, where the $\mathrm{LE}(\cdot)$ is defined as follows: 
$\mathrm{LE}(\cdot)=\mathrm{Up}(\mathrm{CR}(\mathrm{CR}(\cdot)))$,
where $\mathrm{Up}(\cdot)$ is bilinear interpolation and $\mathrm{CR}(\cdot)$ represents the combination of convolution and ReLU. The multi-scale features are then concatenated and passed through a $3\times3$ convolution for feature aggregation to gradually fuse them into $\mathbf Z_1$. Formally, 
\begin{equation}
    \mathbf{Z_4} = F_4^{up}, \quad\text{ and}\quad \mathbf{Z_i} = \mathrm{Conv}_{3\times 3}([\mathbf{Z_{i+1}}, F_i^{up}]), i \in \{3,2,1\}.
\end{equation}
Then we use a simple encoder and decoder to extract detailed information from the input at time $t$ and make predictions under the guidance of condition $\mathbf Z_1$. The encoder and decoder are composed of several convolutional layers and downsampling/upsampling operations, and the time $t$ is incorporated into the convolutional layers using Adaptive Group Normalization~\cite{dhariwal2021diffusion}.

\algrenewcommand\algorithmicindent{0.5em}%
\begin{figure}[t]
  \renewcommand{\arraystretch}{0.5}
  \begin{minipage}[t]{0.495\textwidth}
  \begin{algorithm}[H]
        \caption{Training} \label{alg:training}
    \small
    \begin{algorithmic}[1]
      \Require Image and mask distribution $q(\mathcal I)$, $q(\mathbf x_0|\mathcal I)$
        \vspace{0.35em}
      \Repeat
          \State $t \sim \mathrm{Uniform}(\{1, \dotsc, T\})$
          \State $\mathcal{I} \sim q(\mathcal I)$
          \State $\mathbf{x}_0 \sim q(\mathbf{x}_0|\mathcal I)$
          \State $\boldsymbol \epsilon\sim\mathcal{N}(\bzero,\textbf{I})$
  
          \State $\mathbf{x}_0' = \mathrm{structureCorrupt}(\mathbf{x}_0)$
          \State $\mathbf{x}_t = \alpha_t \mathbf{x}_0' + \sigma_t \boldsymbol \epsilon$
          \State Take gradient descent step on
          \Statex $\qquad {\nabla}_\theta \ell(\mathbf x_0, f(\mathbf x_t, \mathcal I, t))$
      \Until{converged}
      \vspace{0.35em}
    \end{algorithmic}
  \end{algorithm}
  \end{minipage}
  \hfill
  \begin{minipage}[t]{0.495\textwidth}
  \begin{algorithm}[H]
      \caption{Sampling} \label{alg:sampling}
    \small
    \begin{algorithmic}[1]
          \Require Image $\mathcal{I}$
      \State $\mathbf{x}_T \sim \mathcal{N}(\bzero, \textbf{I})$
      \State $\mathbf{x}_p = []$
      \For{$t=T, \dotsc, 1$}
          \State  $\hat{\mathbf{x}}_0 = f_\theta(\mathbf x_t, \mathcal I, t)$
          \State $\mathrm{appendToHistory}(\mathbf{x}_p, \hat{\mathbf{x}}_0) \ \ //\text{Append }\hat{\mathbf{x}}_0\text{ to } \mathbf{x}_p$
          \State $\mathbf{z} \sim \mathcal{N}(\bzero, \textbf{I})$ if $t > 1$, else $\mathbf{z} = \bzero$
          \State $\mathbf x_{t-1} = \frac{\sqrt{\alpha_t}(1 - \bar{\alpha}_{t-1})}{1 - \bar{\alpha}_t} \mathbf{x}_t + \frac{\sqrt{\bar{\alpha}_{t-1}}\beta_t}{1 - \bar{\alpha}_t} \hat{\mathbf{x}}_0 + \sigma_t \mathbf z$
      \EndFor
      \State $\mathbf p = \mathrm{consensusTimeEnsemble}(x_p)$
      \State \textbf{return} $\mathbf p$
    \end{algorithmic}
  \end{algorithm}
  \end{minipage}

  \label{algo:Training-and-Sampling}
\vspace{-3mm}
\end{figure}

\subsection{Training Strategy}
\label{sec:TS}
During Training, we initiate a diffusion process from the GT to the noisy mask and train our model to reverse this process. 
Despite successful training, certain challenges arise. Firstly, the model struggles to recover a clear mask from a low SNR mask according to image features. To address this, an \textbf{SNR-based variance schedule} is adopted to improve the model's performance. The second challenge stems from the inadequacy of Gaussian noise added to the mask during the forward process, and we address it by utilizing \textbf{Structure Corruption} to facilitate the model's ability to learn structure-level denoising. 
These strategies constitute the forward diffusion process, and a comprehensive description of the training procedure can be found in Algorithm \ref{alg:training}.



\normalfont{\normalsize\bfseries \textbullet\ \ SNR-based Variance Schedule:}
It was observed that the presence of masking resulted in excessive SNR in training, which posed a challenge for the model to learn how to recover the mask from low SNR inputs~\cite{chen2022generalist}.
The most common cosine variance schedule was developed for image generation with resolutions of $32^2$ and $64^2$, but it is insufficient to address the issue of adding noise to high-resolution images~\cite{hoogeboom2023simple}, especially for binary masks.
When the SNR is high, the mask is visible to the naked eye, and the model can easily restore the mask without leveraging image features. 

To address this, we adopt an SNR-based variance schedule~\cite{hoogeboom2023simple} to regulate the SNR during training. 
Specifically, according to the definition of SNR~\cite{kingma2021variational}: $\operatorname{SNR}(t)=\bar \alpha_t/(1-\bar \alpha_t)$, 
we designed an SNR schedule, where $ \log \operatorname{SNR}(t)=-2\log(\tan(\frac{\pi t}{2}))$.
Subsequently, we introduce an offset to it, expressed as: $\log \operatorname{SNR}_{\text{shift}}(t)=-2\log(\tan(\frac{\pi t}{2}))+\text{shift}$. By reducing the shift value, fewer easy cases are reversed at the same time step $t$.




\normalfont{\normalsize\bfseries \textbullet\ \ Structure Corruption:}
Existing diffusion models employ pixel-level corruption to generate the noised mask directly from the GT, leading the model to incorrectly assume that the restored contour from the noised mask is always accurate, resulting in an inability to correct the bias. This assumption, however, is not guaranteed to hold during the sampling. To address this issue, we propose Structure Corruption during forward diffusion, where we randomly destroy the contour of the GT and add Gaussian noise. This enables the model to learn how to restore the correct mask from the noised mask, even with bias.





\subsection{Sampling Strategy}
\label{sec:SS}
Our denoising model applies incremental denoising to a sample $\mathbf x_T$ drawn from a standard normal distribution over $T$ steps. This iterative denoising process steadily mitigates the divergence between the predicted mask and the ground truth, culminating in a more precise outcome. 
Although the final denoised mask exhibits clear and well-defined boundaries, we contend that the predictions generated during the denoising process also offer valuable insights. Therefore, we utilize a \textbf{Consensus Time Ensemble} approach, which combines predictions from each sampling step to improve the result's precision and reliability. 
The pseudo-code of the sampling strategy is detailed in Algorithm~\ref{alg:sampling}.


\normalfont{\normalsize\bfseries \textbullet\ \ Consensus Time Ensemble:}
Inspired by the annotation procedure of saliency detection~\cite{fan2020rethinking, zhang2021uncertainty}, we propose a method called Consensus Time Ensemble (CTE) to combine predictions from different sampling steps without incurring additional computational costs. Specifically, for each Sampling stage at time $t$, the denoised image $\hat{\mathbf x}_0$ is denoted as $P_t$. Given multiple predictions $\{P_t\}_{t=1}^T$, the binary masks $\{P^{t}_b\}_{t=1}^T$ are first calculated through adaptive thresholding. 
These predictions $\{P^{t}_b\}_{t=1}^T$ vote on the position of each point to generate a candidate mask, and the probability value of the selected point is the mean of all predictions.
Mathematically, 
\vspace{-1.5mm}
\begin{equation}
    P_{emb} = \left\lfloor\frac{\sum_{t=1}^T P_b^t}{T} + \frac{1}{2}\right\rfloor * \mathrm{mean}(P_t).
\end{equation}

\vspace{-2.5mm}
In addition, our model can sample from the mask distribution, enabling us to make multiple predictions and run an ensemble to obtain a more accurate mask. In practice, we sample the mask thrice and apply our CTE strategy to combine these predictions $\{P_t\}_{t=1}^{3T}$. 
The performance of this approach is presented as "\ourmodel-E" in our evaluation.


\begin{table*}[t!]
  \centering
  \footnotesize
  \small{
    \caption{
      Quantitative results of our method and other state-of-the-art methods on four benchmark datasets. "-P" and "-S" means taking PVTv2 and Swin Transformer as a backbone. The best results are highlighted in \textbf{bold}, and the second-best results are marked with an \underline{underline}.
      }\label{tab:ModelScore}
  }
  \small
  \renewcommand{\arraystretch}{0.9} 
  \setlength\tabcolsep{3.7pt} 
  \resizebox{1.0\textwidth}{!}{
  \begin{tabular}{rcccccccccccc}
  \toprule
    \multirow{2}*{Baseline Models~~~} & \multicolumn{4}{c}{\tabincell{c}{CAMO(250)~\cite{le2019anabranch}}} & \multicolumn{4}{c}{ \tabincell{c}{COD10K(2026) \cite{fan2020camouflaged}}} & \multicolumn{4}{c}{ \tabincell{c}{NC4K(4121) \cite{lv2021simultaneously}}} \\
   \cmidrule(lr){2-5} \cmidrule(lr){6-9} \cmidrule(lr){10-13}
     &$S_\alpha\uparrow$      &$E_\phi\uparrow$     &$F_\beta^w\uparrow$      &$M\downarrow$
               &$S_\alpha\uparrow$      &$E_\phi\uparrow$     &$F_\beta^w\uparrow$      &$M\downarrow$
               &$S_\alpha\uparrow$      &$E_\phi\uparrow$     &$F_\beta^w\uparrow$      &$M\downarrow$
               \\
     \hline         
         
    MirrorNet\textsubscript{2020}~\cite{yan2020mirrornet}
                    &0.741 &0.804 &0.652& 0.100& $\ddagger$ & $\ddagger$ & $\ddagger$ & $\ddagger$  & $\ddagger$ & $\ddagger$ & $\ddagger$& $\ddagger$  \\ 
    PraNet\textsubscript{2020}~\cite{fan2020pranet}
                    & 0.769& 0.833 &0.663& 0.094& 0.789& 0.839& 0.629& 0.045 & 0.822 & 0.876 & 0.724 & 0.059\\
                    
    SINet\textsubscript{2020}~\cite{fan2020camouflaged}
                    &0.751 &0.771& 0.606 &0.100& 0.771& 0.806& 0.551 &0.051     &  0.810 & 0.873 & 0.772 & 0.057 \\  
    
    MGL\textsubscript{2021}~\cite{zhai2021mutual}
    &0.775 &0.847 &0.673 &0.088
    &0.814 &0.865 &0.666 &{0.035} 
    &$\ddagger$ &$\ddagger$ &$\ddagger$ &$\ddagger$\\
    
    PFNet\textsubscript{2021}~\cite{mei2021camouflaged} 
    &0.782 &0.852 &0.695 &0.085
    &0.800 &0.868 &0.660 &0.040 
    &0.829 &0.887 &0.745 &0.053 \\
    
    UGTR\textsubscript{2021}~\cite{yang2021uncertainty} 
    &0.785 &0.859 &0.686 &0.086
    &{0.818} &0.850 &0.667 &{0.035} 
    &$\ddagger$ &$\ddagger$ & $\ddagger$& $\ddagger$ \\
    
    LSR\textsubscript{2021}~\cite{lv2021simultaneously}  
    &0.793 &0.826 &0.725 &0.085 
    &0.793 &0.868 &{0.685} &0.041 
    &0.839 &0.883 &{0.779} &0.053 \\

    PreyNet\textsubscript{2022}~\cite{zhang2022preynet}
    &0.790 &0.842 &0.708 &0.077
    &0.813 &0.891 &0.697 &0.034
    &$\ddagger$ &$\ddagger$ & $\ddagger$& $\ddagger$ \\

    SINet-V2\textsubscript{2022}~\cite{fangtpami2021} 
    &0.820 &0.882 &0.743 &0.070
    &0.815 &0.887 &0.680 &0.037 
    &0.847 &0.903 &0.769 &0.048 \\
    
    ZoomNet\textsubscript{2022}~\cite{pang2022zoom}
    &0.820 &0.892 &0.752 &0.066
    &0.838 &0.888 &0.729 &0.029
    &0.853 &0.896 &0.784 &0.043 \\

    DTINet\textsubscript{2023} ~\cite{liu2022boosting}
    &0.856 &0.916 &0.796 &0.050
    &0.824 &0.896 &0.695 &0.034 
    &0.863 &0.917 &0.792 &0.041 \\
    
    FEDER-R50\textsubscript{2023}~\cite{hecamouflaged}
    &0.807 &0.873 &$\ddagger$ &0.069
    &0.823 &0.900 &$\ddagger$ &0.032
    &0.846 &0.905 &$\ddagger$ &0.045 \\

    FEDER-R2N\textsubscript{2023}~\cite{hecamouflaged}
    &0.836 &0.897 &$\ddagger$ &0.066
    &0.844 &0.911 &$\ddagger$ &0.029
    &0.862 &0.913 &$\ddagger$ &0.042 \\
    
    HitNet\textsubscript{2023}~\cite{hu2023high}
    &0.844 &0.902 &0.801 &0.057
    &0.868 &0.932 &0.798 &0.024 
    &0.870 &0.921 &0.825 &0.039 \\

    FSPNet\textsubscript{2023}~\cite{huang2023feature}
    &0.856 &0.899 &0.799 &0.050
    &0.851 &0.895 &0.735 &0.026
    &0.879 &0.915 &0.816 &0.035 \\

    DGNet \textsubscript{2023}~\cite{ji2023gradient}
    &0.839 &0.901 &0.769 &0.057
    &0.822 &0.896 &0.693 &0.033
    &0.857 &0.911 &0.784 &0.042 \\
    
    DGNet-P\textsubscript{2023}~\cite{ji2023gradient}
    &0.877 &0.930 &0.831 &0.046
    &0.857 &0.922 &0.760 &0.026
    &0.882 &0.931 &0.829 &0.035 \\
    
    CamoFormer-S \textsubscript{2023} ~\cite{yin2023camoformer}
    &0.876 &0.930 &0.832 &\underline{0.043}
    &0.862 &0.931 &0.772 &0.024 
    &0.888 &0.937 &0.840 &0.031 
    \\
    CamoFormer-P \textsubscript{2023} ~\cite{yin2023camoformer}
    &0.872 &0.929 &0.831 &0.046 
    &0.869 &0.932 &0.786 &0.023
    &0.892 &\underline{0.939} &0.847 &\underline{0.030} \\         \hline 
    \ourmodel(Ours) 
    &\underline{0.879} &\textbf{0.940} &\underline{0.854} &\textbf{0.042}
    &\underline{0.880} &\textbf{0.943} &\underline{0.815} &\underline{0.020}
    &\underline{0.892} &\textbf{0.941} &\underline{0.858} &\textbf{0.029} \\
    
    \ourmodel-E(Ours)
    &\textbf{0.880} &\underline{0.939} &\textbf{0.855} &\textbf{0.042}
    &\textbf{0.883} &\underline{0.942} &\textbf{0.819} &\textbf{0.019}
    &\textbf{0.894} &\textbf{0.941} &\textbf{0.859} &\textbf{0.029} \\
  \toprule
  \end{tabular}
 }
\vspace{-3mm}
\end{table*}    
\section{Experiment}
\label{others}
\subsection{Experiment Settings}
\paragraph{Datasets.}
Our \ourmodel\ is evaluated on three widely-used COD datasets: CAMO~\cite{le2019anabranch}, COD10K~\cite{fangtpami2021}, and NC4K~\cite{lv2021simultaneously}. The CAMO dataset comprises 2,500 images with 2,000 images for training and 500 images for testing. 
COD10K contains 5,066 camouflaged, 3,000 background, and 1,934 non-camouflaged images.
NC4K is a large-scale COD dataset consisting of 4,121 images and is often used to evaluate the generalization ability of models. 
\vspace{-8pt}
\paragraph{Evaluation Metrics.} 
In order to evaluate the performance of our proposed method, we adopt four commonly used metrics: mean absolute error (MAE)~\cite{perazzi2012}, S-measure ($S_\alpha$)~\cite{fan2017structure}, weighted F-measure ($F^\omega_\beta$)~\cite{margolin2014evaluate}, and mean E-measure ($E_\phi$)~\cite{fan2021cognitive}. 
\vspace{-8pt}
\paragraph{Implementation Details.} 
We implement our \ourmodel\ based on PyTorch using a single NVIDIA A100 with 40GB memory for both training and inference. For training efficiently, we initiate the training process with image sizes of 352 for 150 epochs, followed by fine-tuning for an additional 20 epochs with image sizes of 384. For optimization, the AdamW~\cite{loshchilov2017decoupled} optimizer was utilized along with a batch size set to 32. To adjust the learning rate, we implemented the cosine strategy with an initial learning rate of 0.001.
\subsection{Comparisons with State-of-the-Arts Methods}
In this study, we conducted a comprehensive comparison between the proposed \ourmodel\ and 19 state-of-the-art methods. 
Since our proposed method samples different predictions each time, we evaluated it three times with different seeds and report the average of the metrics for \ourmodel. Additionally, we introduced \ourmodel-E, which ensembles three predictions sampled from the mask distribution to improve the reliability of the model through our CTE strategy.
Notably, our default configuration employs $T=10$ for the sampling process.
\vspace{-7pt}
\paragraph{Quantitative Comparison.}
Tab.~\ref{tab:ModelScore} summarizes the quantitative results of our proposed method against 19 competitors on three challenging COD benchmark datasets. The results show that our \ourmodel\ outperforms all other models on these datasets. 
Moreover, our approach outperforms models like CamoFormer-P~\cite{yin2023camoformer} and DGNet-P~\cite{ji2023gradient} that are also based on PVTv2, showcasing the efficacy of our methodology.
Specifically, our model dramatically reduces the MAE error by $17.4\%$ and increases $F_\beta^w$ by $4.2\%$ compared to the second-best performer, CamoFormer-P~\cite{yin2023camoformer}.
The superiority in performance benefits from the iterative denoising process of diffusion models.

\begin{figure}
  \centering
  \small
  \includegraphics[width=1.0 \textwidth,height=0.7\textwidth, trim=20 60 20 0,clip]{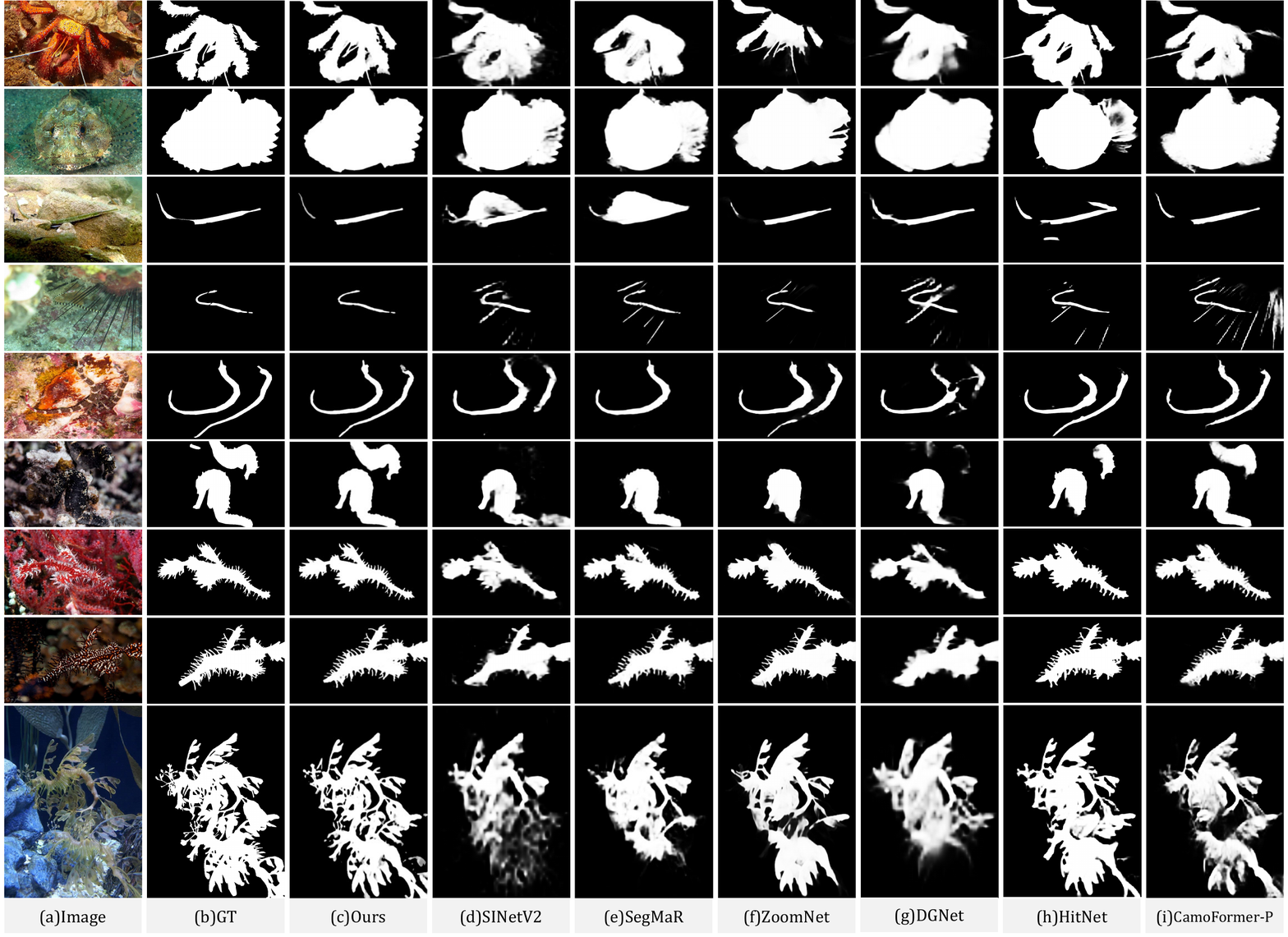}
    \caption{Visual comparisons with recent SOTA models in typical and challenging scenarios. Our \ourmodel\ can generate more accurate predictions, showcasing exceptional performance in intricate details and possessing well-defined contours.}
  \label{fig:sotaVisualCompare}
  \vspace{-3mm}
\end{figure}

\vspace{-7pt}
\paragraph{Visual Comparison.}
Fig.~\ref{fig:sotaVisualCompare} presents the visual comparisons of \ourmodel\ with several recent SOTA COD models. The examples illustrated in the first six rows exhibit typical scenarios, including objects of large or small sizes and multiple objects. Our proposed \ourmodel\ performs well in these cases. The examples in the seventh to ninth rows contain complex topological structures and detailed edges, which pose significant challenges to current COD models. However, our \ourmodel\ is capable of accurately segmenting the clear boundaries of these examples, due to our iterative denoising paradigm. 
Notably, HitNet~\cite{hu2023high} also performs well on these examples, but it utilizes high-resolution images of $704^2$ as input, whereas \ourmodel\ achieves better results using only images of $384^2$ resolution.
\begin{table}[htbp]
  \footnotesize
  \centering
  \setlength\tabcolsep{4pt}
  \small
  \caption{Quantitative results on different models and learning strategies.
  }\label{tab:module_ablation}
  \begin{subtable}{0.49\textwidth}
    \centering
      \begin{tabular}{lcccc}
      \toprule
      \multirow{2}*{Model Setting~~~ }&\multicolumn{4}{c}{\tabincell{c}{COD10K~\cite{fan2020camouflaged}}} \\
      \cmidrule(lr){2-5}         
      &$S_\alpha\uparrow$      &$E_\phi\uparrow$     &$F_\beta^w\uparrow$      &$M\downarrow$  \\ \hline 
      \ourmodel&\textbf{0.880} &\textbf{0.943} &\textbf{0.815} &\textbf{0.020} \\
      Baseline &0.866 &0.928 &0.775 &0.024 \\ 
      w/o ZOE &0.846 &0.918 &0.754 &0.027 \\
      w/o ATCN &0.869 &0.936 &0.795 &0.022 \\
      \bottomrule
    \end{tabular}
    \caption{Comparison of Model Settings.}
    \label{table:modelSettingCompare}
  \end{subtable}
  \hfill
  \begin{subtable}{0.49\textwidth}
    \centering
    \renewcommand{\arraystretch}{1.0}
      \begin{tabular}{lcccc}
      \toprule
      \multirow{2}*{Learning Strategy~~~ }&\multicolumn{4}{c}{\tabincell{c}{COD10K~\cite{fan2020camouflaged}}}  \\
      \cmidrule(lr){2-5}
       &$S_\alpha\uparrow$      &$E_\phi\uparrow$     &$F_\beta^w\uparrow$      &$M\downarrow$\\ \hline
      \ourmodel&\textbf{0.880} &\textbf{0.943} &\textbf{0.815} &\textbf{0.020} \\
      w/o SNR-adj &0.855 &0.922 &0.767 &0.024 \\
      w/o SC &0.868 & 0.931 & 0.789 & 0.022 \\ 
      w/o CTE & 0.877 & 0.941 & 0.803 & 0.021  \\
      \bottomrule
    \end{tabular}
    \caption{Comparison of Learning Strategies.}
    \label{table:trainingSettingCompare}
  \end{subtable}   
  \vspace{-7mm}
\end{table}
\subsection{Ablation Studies}
\paragraph{Ablation of Our Model Setting.}
To assess the effectiveness of our method, we compare it with a PVTv2-based baseline approach that follows the segmentation paradigm, as presented in the first two rows of Tab.~\ref{table:modelSettingCompare}.
Specifically, the baseline decoder contains the LE module~\cite{wang2022stepwise}, concatenation, and upsampling operation.
Our results show that the integration of the diffusion framework significantly improves the performance of the proposed method. 

Additionally, we conduct an ablation study on the individual components of \ourmodel, ATCN, and ZOE, as reported in Tab.\ref{table:modelSettingCompare}. By utilizing ATCN, our model can gradually focus on the object, as described in Sec.\ref{sec:conclusion}, resulting in more accurate predictions. Furthermore, our findings suggest that the absence of ZOE leads to inferior performance compared to the baseline, primarily because the random initialization of the convolution layer fails to add positional information to the embeddings, resulting in slower convergence.

\paragraph{Ablation of Our Learning Strategy.}
We evaluate the effectiveness of the Structure Corruption and SNR-based variance schedule employed in our proposed \ourmodel. Tab.~\ref{table:trainingSettingCompare} presents the experimental results obtained by comparing our model trained with and without these strategies. Specifically, the "w/o SNR-adj" configuration denotes the addition of Gaussian noise based on the cosine variance schedule, while the "w/o SC" configuration employs only pixel-level corruption during the forward process, without our proposed Structure Corruption. Our results demonstrate that the integration of these two training strategies improves the performance of our model.

Furthermore, we investigate the effects of the CTE method, which combines predictions from multiple steps to generate more robust predictions. 
Rows 1, and 4 of Tab.~\ref{table:trainingSettingCompare} show the effects of CTE, 
where "w/o CTE" denotes utilizing $\mathbf{x}_0$ as the prediction. 
Our experimental results indicate that the CTE strategy significantly improves the quality of the generated masks.
\paragraph{Ablation of Hyperparameters.}
\begin{figure}
\centering
\small
\begin{minipage}[t]{0.49\textwidth}
  \centering
  \small
  \includegraphics[width=1.0 \textwidth, trim=5 85 0 230,clip]{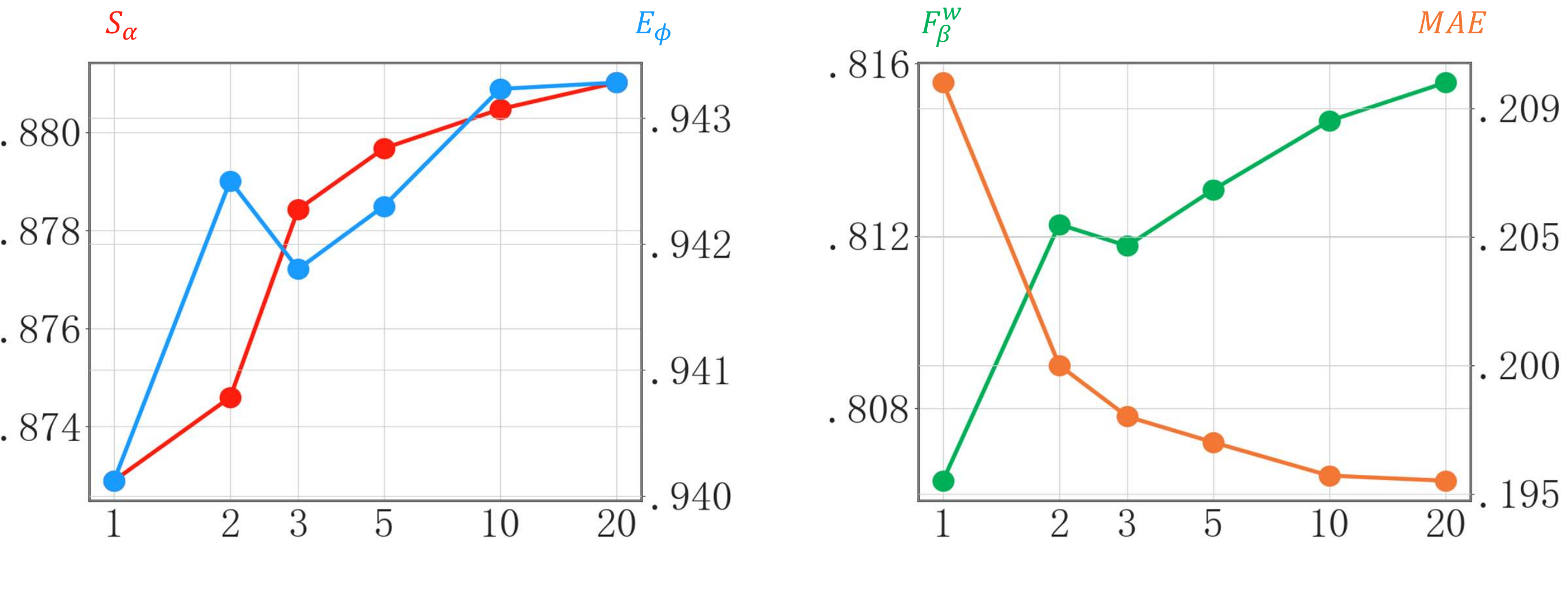}\vspace{-2mm}
  \caption{Performance on different sample steps.}
  \label{fig:hyper-parameterSetting:step}
\end{minipage}
\tikz{\draw[-,gray, densely dashed, thick](0,-1.35) -- (0,1.0);}
\begin{minipage}[t]{0.49\textwidth}
  \centering
  \small
  \includegraphics[width=1.0 \textwidth, trim=2 90 5 230,clip]{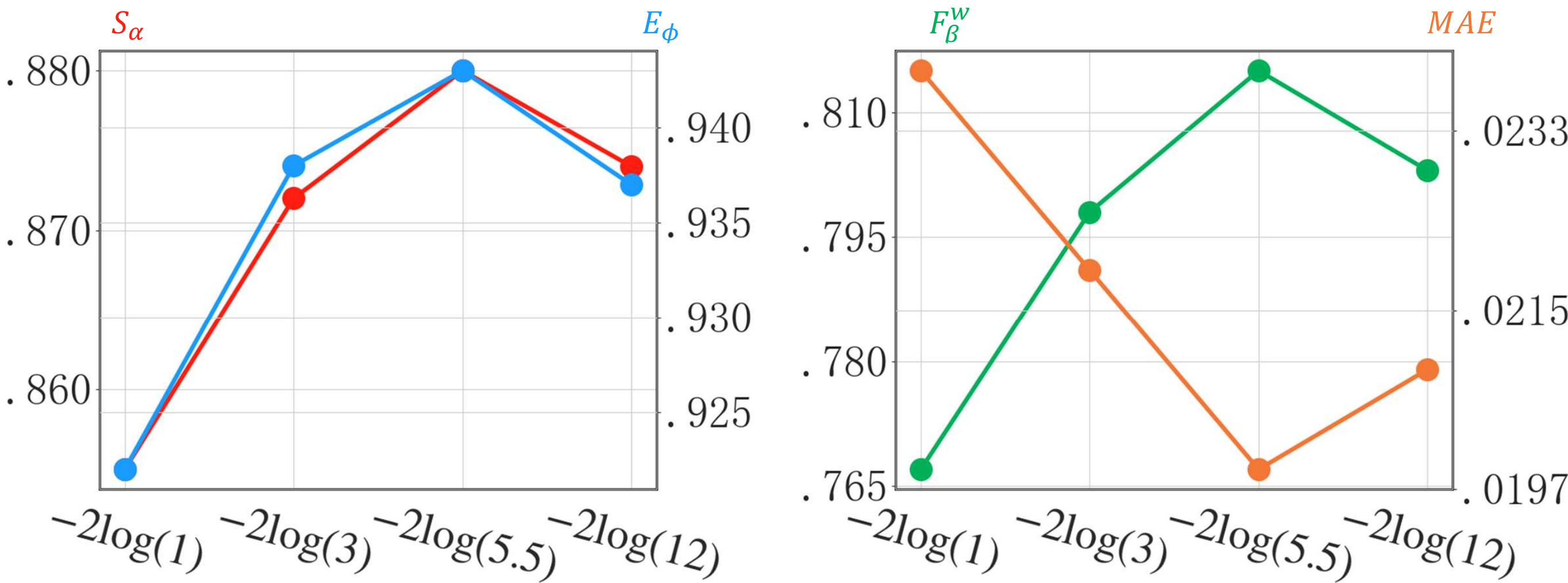}\vspace{-2mm}
  \caption{Performance on different SNR Shift.}
  \label{fig:hyper-parameterSetting:shift}
\end{minipage}
\vspace{-3mm}
\end{figure}
Fig.~\ref{fig:hyper-parameterSetting:step} presents the performance of the COD10k dataset with varying numbers of sampling steps. The result reveals that the model's performance improves as the number of sampling steps increases, but it also leads to longer inference time. We set the sampling steps to $10$ for a trade-off between performance and computational cost.  
Moreover, the impact of SNR Shift on model performance is crucial, as presented in Fig.~\ref{fig:hyper-parameterSetting:shift}. It is worth noting that shifting down the shift value reduces the amount of mask information preserved in $\mathbf x_t$, which in turn affects the training difficulty. The optimal performance is observed when the shift is equal to $-2\log(5.5)$, which provides an appropriate level of difficulty for training.

\subsection{Analysis}
\begin{figure}[t!]
\small 
\begin{subfigure}[t]{0.46\textwidth}
  \centering
  \small
  \includegraphics[width=1.0 \textwidth, trim=20 100 20 80,clip]{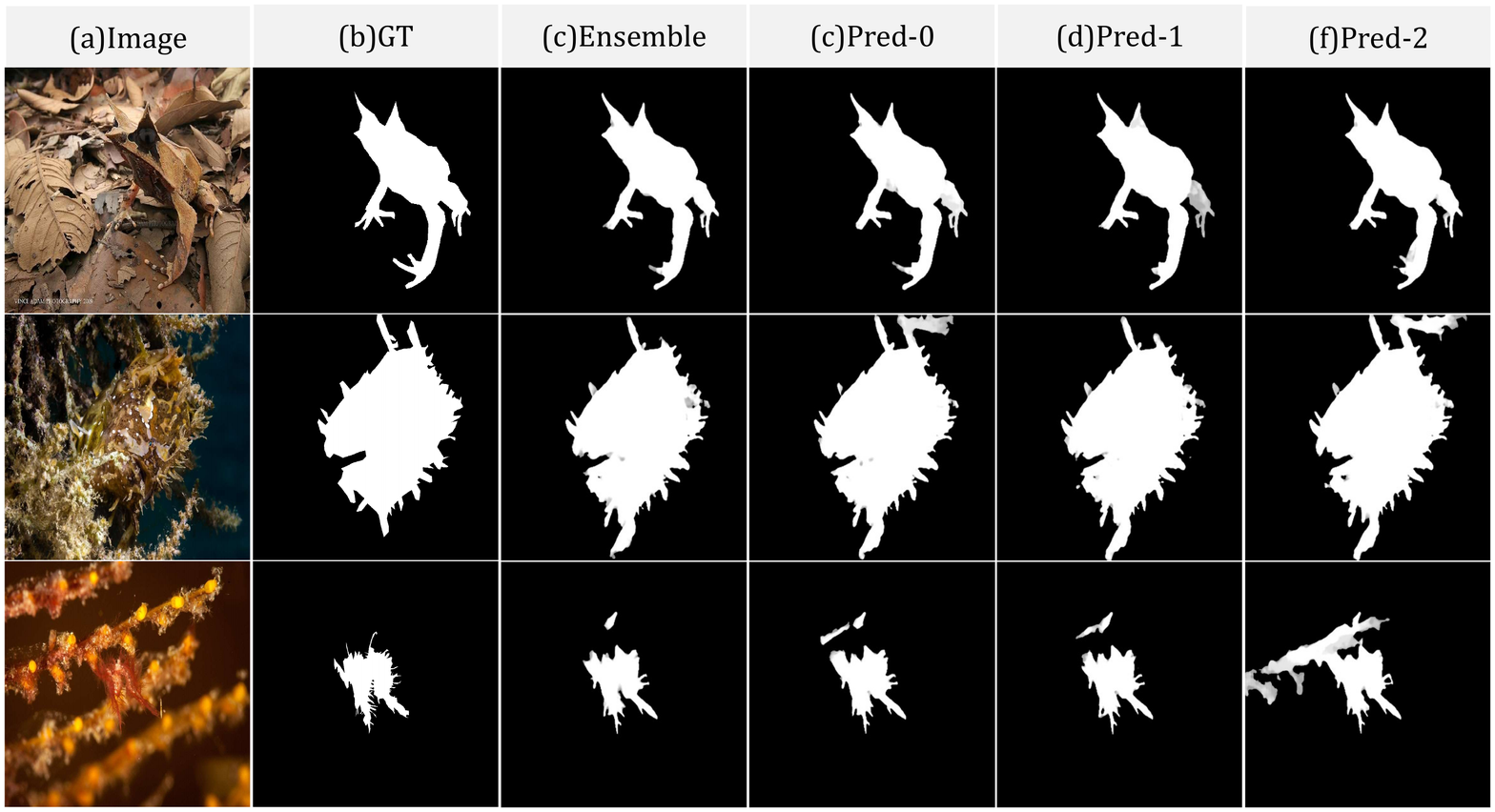}
  \vspace{-6.5mm}
  \caption{Ensemble of multiple predictions.}
  \label{fig:ensembleCompare}
\end{subfigure}
\begin{subfigure}[t]{0.52\textwidth}
  \centering
  \includegraphics[width=1.0 \textwidth, trim=0 150 120 120,clip]{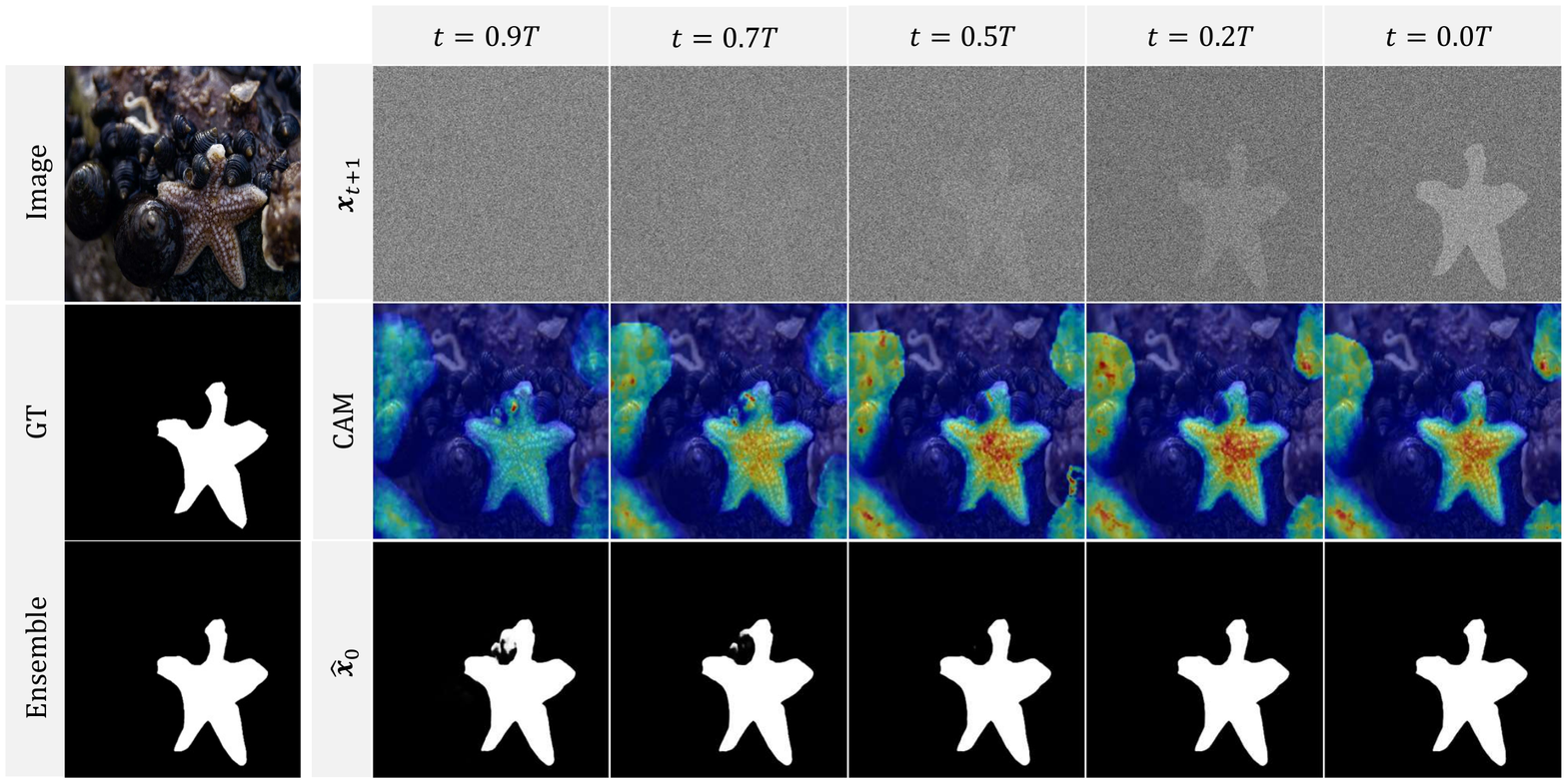}
  \small
  \vspace{-6.5mm}
  \caption{Mask predictions and feature maps at different steps.}
  \label{fig:inferTimeCompare}
\end{subfigure}
\small
\vspace{-2mm}
\caption{(a) presents the performance visualization of \ourmodel-E on sample images from the COD10K dataset. 
(b) illustrates the mask predictions and feature maps at different sample steps. 
}
\vspace{-3mm}
\end{figure}
As illustrated in Fig.~\ref{fig:ensembleCompare}, our proposed \ourmodel-E leverages the stochastic sampling process of diffusion to generate a distribution of segmentation masks, resulting in multiple predictions denoted as "Pred-0" through "Pred-2". These predictions display discrepancies in certain regions, indicating the model's uncertainty in those areas. To enhance the robustness of our final prediction, we utilize CTE to combine them into a single mask labeled as "Ensemble".

To demonstrate our model's capacity to reduce noise and progressively concentrate on intricate details, we present Fig.~\ref{fig:inferTimeCompare}, which depicts the prediction results at various sampling stages.
Initially, the model produces a coarse mask with a high degree of uncertainty in specific regions where borders are unclear. However, as the number of sampling steps increases, the model progressively concentrates on the concealed object and refines the mask, establishing deterministic boundaries based on the subtle details of the foreground.
This iterative refinement is facilitated by the incorporation of diffusion models, which substantially contribute to the improvement of the model's performance.

\section{Conclusion}
\label{sec:conclusion}
In this study, we present a diffusion-based COD model, named \ourmodel, which is composed of ATCN and DN. Our model treats mask prediction as a distribution estimation problem and utilizes the diffusion framework to iteratively reduce the noise of the mask. 
Concerning the learning strategies, we devise a novel ensemble method to achieve a robust mask and adapt forward diffusion techniques specifically for the COD task to ensure efficient training. 
Our experiments demonstrate the superiority of our method over existing SOTA methods.

{\small 
\bibliographystyle{IEEEtranN}
\bibliography{main}
}
\end{document}